\documentclass{article} 
\usepackage{nips15submit_e,times}
\usepackage{hyperref}
\usepackage{url}
\usepackage{amsthm}

\usepackage{amsmath}
\usepackage{amssymb}
\usepackage{algorithm}
\usepackage{algorithmic}
\usepackage[square,sort,comma,numbers]{natbib}

\newcommand{\ignore}[1]{}
\newcommand{\maxmu}{\mu^{*}}

\newcommand{\KK}{K}

\newcommand{\Indi}[1]{{I\left(#1\right)}}

\newcommand{\G}{L}

\newcommand{\UB}{\epsilon^{UB}}

\newtheorem{assumption}{Assumption}
\newtheorem{theorem}{Theorem}
\newtheorem*{proof*}{Proof}
\newtheorem{example}{Example}
\newtheorem{corollary}{Corollary}
\newcommand{\CA}{A}
\newcommand{\GME}{L^{ME}}

\title{The Max $K$-Armed Bandit: A PAC Lower Bound and tighter Algorithms}

\author{
Yahel David and Nahum Shimkin\\
Department of Electrical Engineering\\
Technion - Israel Institute of Technology\\
Haifa 32000, Israel
}

\nipsfinalcopy 

\begin{document}

\maketitle

\begin{abstract}
We consider the Max $K$-Armed Bandit problem, where a learning agent is faced with several sources (arms) of items (rewards), and interested in finding the
best item overall. At each time step the agent chooses an arm, and obtains a random real valued reward. The rewards of each arm are assumed to be i.i.d., with an unknown probability distribution that generally differs among the arms.
Under the PAC framework, we provide lower bounds on the sample complexity of any
$(\epsilon,\delta)$-correct algorithm, and propose algorithms that attain this bound
up to logarithmic factors.
We compare the performance of this multi-arm algorithms to the variant in which the arms
are not distinguishable by the agent and are chosen randomly at each stage.
Interestingly, when the maximal rewards of the arms happen to be similar, the latter
approach may provide better performance.
\end{abstract}

\section{Introduction}
In the classic stochastic multi-armed bandit (MAB) problem the learning agent faces a set $\KK$ of stochastic arms, and wishes to maximize its cumulative reward (in the regret formulation), or find the arm with the highest mean reward (the pure exploration problem). This model has been studied extensively in the statistical and learning literature, see for example \cite{bubeck2012regret} for a comprehensive survey.

We consider a variant of the MAB problem called the Max $K$-Armed Bandit problem (Max-Bandit for short).
In this variant, the objective is to obtain a sample with the highest possible reward (namely, the highest value in the support of the probability distribution of any arm).
More precisely, considering the PAC setting, the objective is to return an $(\epsilon,\delta)$-correct sample,
namely a sample which its reward value is $\epsilon$-close to the overall best possible reward with a probability larger than $1-\delta$. In addition, we wish to minimize the sample complexity, namely the expected number of samples observed by the learning algorithm before it terminates.

For the classical MAB problem, algorithms that find the best arm (in terms of its expected reward) in the PAC sense were presented in \cite{Dar,audibert2010bestArmIdentification,BestArmIdentification_AUnifiedApproach}, and
lower bounds on the sample complexity were presented in \cite{PAC_lower} and \cite{audibert2010bestArmIdentification}. The essential difference with respect to this work is in the objective, which is to find an $(\epsilon,\delta)$-correct sample in our case.
The scenario considered in Max-Bandit model is relevant when a single best item needs to be selected from
among several (large) clustered sets of items, with each set represented as a single arm.
These sets may represent parts that come from different manufacturers
or produced by different processes, job candidates that are referred by different employment agencies, finding the best match to certain genetic characteristics in different populations,  or choosing the best channel among different frequency bands in a cognitive radio wireless network.

The Max-Bandit problem was apparently first proposed in \cite{cicirello2005max}. For reward distribution functions in a specific family, an algorithm with an upper bound on the sample complexity that increases as $\frac{-\ln(\delta)}{\epsilon^{2}}$ was provided in \cite{streeter2006asymptotically}.
For the case of discrete rewards, another algorithm was presented in \cite{streeter2006simple}, without performance analysis.
Later, a similar model in which the objective is to maximize the expected value of the largest sampled reward for a given number of samples ($n$) was studied in \cite{Extreme_bandits}. In that work the attained best reward is compared with the expected reward obtained by an oracle that samples the best arm $n$ time. An algorithm is suggested and shown to secure an upper bound of order $n^{-\alpha}$ on that difference, where $\alpha<1$ is determined by the properties of the distribution functions and decreases as they are further away from a specific functions family.

Our basic assumption in the present paper is that a known lower bound is available on the
tail distributions,
namely on the probability that the reward of each given arm will be close to its maximum.
A special case is when the probability densities near the maximum are larger than a given value, but we
consider more general function classes.
Under that assumption, we provide an algorithm for which the sample complexity increases as at most $\frac{-\ln(\delta)-\ln(\epsilon)}{\epsilon}$. This provides an improvement by a factor of $\epsilon^{-1}$ over the result of \cite{streeter2006asymptotically}, which was obtained for a more specific model.
To compare with the result in \cite{Extreme_bandits}, we observe that with a choice of $\delta=\frac{1}{n^2}$ in our algorithm, we obtain that the expected shortfall of the largest sample with respect to the maximal reward possible is at most of order $O(\frac{\ln(n)}{n})$ (as compared to $O(n^{-\alpha})$ with $\alpha<1$).
Furthermore, we provide a lower bound on the sample complexity of every $(\epsilon,\delta)$-correct algorithm,
which holds when several arms posses maximal rewards that are close to that of the best arm.
This lower bound is shown to coincide, up to a logarithmic term, with the upper bound derived for the proposed algorithm.

A basic feature of the Max-Bandit problem (and the associated algorithms) is the goal of quickly
focusing on the best arm (in term of maximal reward), and sampling from that arm as much as possible.
It should be of interest to compare the obtained results with the alternative approach, which ignores the
distinction between arms, and simply draws a sample from a random arm (say, with uniform probabilities) at each round. This can be interpreted as mixing the items associated with each arm
before sampling; we accordingly refer to this variant as the unified-arm problem.
This problem actually coincides with the so-called infinitely-many armed bandit model studied in \cite{berry1997bandit,O._Teytaud,infinitely_many_arms_monus,NIPS_Mortal_Multi_Armed_Bandits,NIPS_Two_Target_Algorithms_for_Inf_Armed_Bandit},
for the specific case of deterministic arms studied in \cite{ECML}.
The conclusion about weather to apply the multi-arm approach or the unified-arm approach is inconclusive. However, as a rule of thumb, when the maximal possible rewards of many arms are far from the optimal, the multi-arm approach has better performance.

The paper proceeds as follows. In the next section we present our model. In Section \ref{sec:lower} we provide a lower bound on the sample complexity of every $(\epsilon,\delta)$-correct algorithm. In Section \ref{sec:alg} we present two $(\epsilon,\delta)$-correct algorithms, and we provide an upper bound on the sample complexity of one of them. The first algorithm is simple and its bound has the same order as the lower bound up to a logarithmic term in $\frac{|\KK|}{\epsilon}$ (where $|\KK|$ stands for the number of arms), the second algorithm is more complicated and we believe that its bound is larger by up to a double logarithmic term in $\frac{|\KK|}{\epsilon}$ than the lower bound. In Section \ref{sec:comp}, we consider for comparison the unified-arm case. In Section \ref{conc} we close the paper
by some concluding remarks. Certain proofs are differed to the Appendix due to space limitations.

\section{Model Definition}\label{sec:model}

We consider a finite set of arms, denoted by $\KK$. At each stage $t=1,2,\dots$ the learning agent chooses an arm $k\in\KK$, and a real valued reward
is obtained from that arm.
The rewards obtained from each arm $k$ are independent and identically distributed, with a distribution function (CDF) $F_{k}(\mu)$, $\mu\in\mathbb{R}$. We denote the maximal possible reward of each arm by $\mu^{*}_{k}=\inf_{\mu\in \mathbb{R}}\{\mu|F_{k}(\mu)=1\}$, assumed finite, and the maximal reward among all arms by $\maxmu=\max_{k\in\KK} \mu^{*}_{k}$.

Throughout the paper, we shall make the following assumption.
\begin{assumption} \label{assumption:1}
	There exist known constants  $\CA>0$, $\beta\geq0$ and $\epsilon_{0}>0$ such that, for every $k\in K$ and $0<\epsilon\leq\epsilon_{0}$, it holds that $$P\left(\boldsymbol{\mu}_{k}>\mu^{*}_{k}-\epsilon\right)\geq\CA\epsilon^{\beta}\,,$$ where $\boldsymbol{\mu}_{k}$ stands for a random variable with distribution $F_{k}$.
\end{assumption}
The bound in the above assumption can also be expressed as $1-F_{k}(\mu)\geq\CA\left(\mu^{*}_{k}-\mu\right)^{\beta}$. This condition required $\boldsymbol{\mu}_{k}$ to have a certain mass near its maximal reward. Note that the specific case of $\beta=1$ is satisfied
if the densities $F'_k$ are lower bounded by a constant $A$. Values of $\beta<1$ accommodate leaner tales.

The upper bound on the CDF $F_{k}$ ensures that for each arm, an $\epsilon$-optimal reward can be observed by a finite number of samples. The bound in the above assumption is similar to those assumed in \cite{infinitely_many_arms_monus} and \cite{ECML}.

An algorithm for the Max-Bandit model samples an arm at each time step, based on the observed history so far (i.e., the previously selected arms and observed rewards). We require the algorithm to terminate after a random number $T$ of samples, which is finite with probability 1, and return a reward $V$
which is the maximal reward observed over the entire period.
An algorithm is said to be \emph{$(\epsilon,\delta)$-correct} if
\begin{equation*}
	P\left(V>\maxmu-\epsilon\right)>1-\delta \,.
\end{equation*}
The expected number of samples $E[T]$ taken by the algorithm is the \emph{sample complexity}, which we wish to minimize.

\section{A Lower Bound}\label{sec:lower}

Before turning to our proposed algorithm, we provide a lower bound on the sample complexity of any $(\epsilon,\delta)$-correct algorithm.
The bounds holds under Assumption \ref{assumption:1} when $\beta\leq 1$. The case of $\beta>1$ is more complicated for analysis and it still unclear whether our lower bound  holds for this case.\ignore{
\begin{assumption}\label{assum2}
	For a given $\epsilon\in(0,\epsilon_{0})$, for every arm $k\in\KK$ for which $\mu^{*}_{k}\geq\maxmu+\epsilon-\epsilon_{0}$, it holds that
	\begin{equation}
		P(\boldsymbol{\mu}_{k}>\mu^{*}_{k}-\Delta_{k}-\epsilon')\geq\CA\left(2\Delta_{k}+\epsilon'\right)^{\beta}\,,
	\end{equation}
	for every $\epsilon'\in[0,\epsilon_{0}-2\Delta_{k}]$, where $\Delta_{k}=\maxmu-\mu^{*}_{k}+\epsilon\leq\frac{\epsilon_{0}}{2}$.
\end{assumption}

It may be verified that if Assumption \ref{assumption:1} holds with a constant $\CA'=\CA2^{\beta}$ (and $\beta>1$),
then Assumption \ref{assum2} holds with the smaller constant $\CA$. In fact we conjecture that Assumption \ref{assum2} is not necessary
for the following lower bound, but a proof is not available at the moment.}

The following result specifies the lower bound of this section.
\begin{theorem}\label{thm:lower:bound}
	Suppose $\epsilon_{0}\leq\left(4\CA\right)^{-1/\beta}$, $\beta\leq1$ and let $\epsilon\in(0,\epsilon_{0})$ and $\delta\in(0,1)$.
	Let $k^*$ denote some optimal arm, such that $\mu^{*}_{k^*} = \maxmu$.
	Then, under Assumption \ref{assumption:1}, for every $(\epsilon,\delta)$-correct algorithm, it holds that
	\begin{equation}
		E[T]\geq \sum_{k\in \KK\setminus \{k^*\}}\frac{1}{8\CA\left(\min\left(\epsilon_{0},\epsilon+\maxmu-\mu^{*}_{k}\right)\right)^{\beta}}\ln\left(\frac{3}{16\delta}\right).
	\end{equation}
\end{theorem}

This lower bound can be interpreted as summing over the minimal number of times that each arm, other than the 
optimal arm $k^*$, needs to be sampled. 
It is important to observe that if there are several optimal arms, only one of them is excluded from the summation.
Indeed, the bound is most effective when there are several optimal (or near-optimal) arms, as the denominator of the summand is larger for such arms. This may appear surprising at first, as more sources of good rewards are available;
however, when there is a single arm that is strictly better than the others it can be quickly singled out,
while if many arms have nearly optimal rewards, more samples are "waisted" on determining which arm is best.

The proof of Theorem \ref{thm:lower:bound} is provided in Appendix A and proceeds by showing that if an algorithm is $(\epsilon,\delta)$-correct and its sample complexity is lower than a certain threshold for some set of reward distributions, then this algorithm cannot be $(\epsilon,\delta)$-correct for some related reward distributions.

\section{Algorithms}\label{sec:alg}
Here we provide two $(\epsilon,\delta)$-correct algorithms. The first algorithm is based on sampling the arm which has the highest upper confidence bound on its maximal reward at each time step and the second algorithm is based on arms elimination.

\subsection{Maximal Confidence Bound}
The algorithm starts by sampling a certain number of times from each arm. Then, it repeatedly calculates an index for each arm which can be interpreted as a certain upper bound on the maximal reward of this arm, and samples once from the arm with the largest index. The algorithm terminates when the number of samples from the arm with the largest index is above a certain threshold. This idea is similar to that in the UCB1 Algorithm provided in \cite{Auer}.
\begin{algorithm}[tb]
	\caption{Maximal Confidence Bound (Max-CB) Algorithm}
	\label{alg:example}
	\begin{algorithmic}[1]
		\STATE {\bfseries Input:} Model parameters $\epsilon_{0}>0$, $\CA>0$ and $\beta\geq0$, constants $\delta>0$ and $\epsilon>0$.\\ Define $\G=6\ln\left(|\KK|\left(1+\frac{-\ln(\delta)}{\CA\epsilon^{\beta}}\right)\right)$.
		\STATE {\bfseries Initialization:} Counters $C(k)=N_{0}$, $k\in K$, where $N_{0}=\lfloor\frac{L-\ln(\delta)}{\CA\epsilon^{\beta}_{0}}\rfloor+1$.
		\STATE Sample $N_{0}$ times from each arm.\label{begin:mu}
		\STATE Compute $Y^{k}_{C(k)}=V^{k}_{C(k)}+\UB(C(k))$ and set $k^{*} \in \arg\max_{k\in\KK}Y^{k}_{C(k)}$ (with tie broken arbitrary),
		where $V^{k}_{C(k)}$ is the largest reward observed so far from arm $k$ and $$\UB(C(k))=\left(\frac{\G-\ln(\delta)}{\CA C(k)}\right)^{1/\beta}\,.$$\label{cond1}
		\STATE If $\UB(C\left(k^{*}\right))<\epsilon$, stop and return the largest sampled reward.\\
		Else, sample once from arm $k^{*}$, set $C(k^{*})=C(k^{*})+1$ and return to step \ref{cond1}.\label{cond2}
	\end{algorithmic}
\end{algorithm}
\begin{theorem}\label{thm:alg1}
Under Assumption \ref{assumption:1}, for $\G\geq10$, Algorithm \ref{alg:example} is $(\epsilon,\delta)$-correct with a sample complexity of
\begin{equation*}
E[T]\leq
\sum_{k\in\KK}\frac{\G-\ln(\delta)}{\CA \left(\max\left(\epsilon,\maxmu-\mu^{*}_{k}\right)\right)^{\beta}}+|\KK|N_{0},
\end{equation*}
where $N_{0}=\lfloor\frac{L-\ln(\delta)}{\CA\epsilon^{\beta}_{0}}\rfloor+1$ and $\G=6\ln\left(|\KK|\left(1+\frac{-\ln(\delta)}{\CA\epsilon^{\beta}}\right)\right)$ as defined in the algorithm.
\end{theorem}

In the following corollary we present the ratio between the lower bound presented in Theorem \ref{thm:lower:bound} to the upper bound in Theorem \ref{thm:alg1}.
\begin{corollary}
If there are more than one arm for which $\mu^{*}_{k}\in[\maxmu-\epsilon,\maxmu]$, then the upper bound on the sample complexity is of the same order as the lower bound in Theorem \ref{thm:lower:bound}, up to a logarithmic factor in $\frac{|\KK|}{\epsilon}$.
\end{corollary}
\begin{proof*}
For every $k\in\KK$ it follows that
$$\Theta^{1}_{k}\triangleq\frac{1+2^{\beta}}{\left(\min\left(\epsilon_{0},\epsilon+\maxmu-\mu^{*}_{k}\right)\right)^{\beta}}\geq\frac{2^{\beta}}{\left(\epsilon+\maxmu-\mu^{*}_{k}\right)^{\beta}}+\frac{1}{\epsilon_{0}^{\beta}}\geq\frac{1}{\left(\max\left(\epsilon,\maxmu-\mu^{*}_{k}\right)\right)^{\beta}}+\frac{1}{\epsilon_{0}^{\beta}}\triangleq\Theta^{2}_{k}\,,$$
and for every two arms $k'$ and $k^{*}$ for which $\mu^{*}_{k'}\in[\maxmu-\epsilon,\maxmu]$ and $\mu^{*}_{k^{*}}=\maxmu$ it is obtained that 
\begin{equation}
\Theta^{1}_{k'}\geq2^{-\beta}\Theta^{1}_{k^{*}}\,.
\end{equation}
In addition, the lower bound is of the same order as
\begin{equation}\label{cor:lower:uper}
-\ln(\delta)\sum_{k\in \KK\setminus \{k^*\}}\Theta^{1}_{k}\,,
\end{equation}
the upper bound is of the same order as $$\left(\G-\ln(\delta)\right)\sum_{k\in \KK}\Theta^{2}_{k}\,,$$
Therefore, the upper bound in Theorem \ref{thm:alg1} is of the same order of the lower bound in Theorem \ref{thm:lower:bound} up to an order of $\frac{\G-\ln(\delta)}{-\ln(\delta)}$, which is logarithmic in $\frac{|K|}{\epsilon}$.

\qed
\end{proof*}

To establish Theorem \ref{thm:alg1}, we first bound the probability of the event under which the upper bound of the best arm is below the maximal reward. Then, we bound the largest number of samples after which the algorithm terminates under the assumption that the upper bound of the best arm is above the maximal reward.

\begin{proof*}[Theorem \ref{thm:alg1}]
We denote the time step of the algorithm by $t$, and the value of the counter $C(k)$ at time step $t$ by $C^{t}(k)$. Recall that $T$ stands for the random final time step. By the condition in step \ref{cond2} of the algorithm, for every arm $k\in\KK$, it follows that,
\begin{equation}\label{C_bound}
C^{T}(k)\leq \lfloor\frac{\G-\ln(\delta)}{\CA \epsilon^{\beta}}\rfloor+1.
\end{equation}
Note that by the fact that for $x\geq6$ it follows that $\frac{d6\ln(x)}{dx}\leq1$, and by the fact that for $x_{0}=\exp\left(1\frac{2}{3}\right)$ it follows that $x_{0}>6\ln(x_{0})=10$ it is obtained that
\begin{equation*}
L'\triangleq|\KK|\left(\frac{-\ln(\delta)}{\CA \epsilon^{\beta}} +1\right)>6\ln\left(|\KK|\left(\frac{-\ln(\delta)}{\CA \epsilon^{\beta}} +1\right)\right)=\G,
\end{equation*}
for $L\geq10$.
So, by the fact that $T=\sum_{k\in\KK}C^{T}(i)$, for $\G\geq10$ it follows that
\begin{equation}\label{time:bound}
T\leq |\KK|\left(\frac{\G-\ln(\delta)}{\CA \epsilon^{\beta}} +1\right) < |\KK|\left(\frac{L'-\ln(\delta)}{\CA \epsilon^{\beta}} +1\right) \leq L'^{2}=e^{\frac{\G}{3}}.
\end{equation}
Now, we begin with proving the $(\epsilon,\delta)$-correctness property of the algorithm.
Recall that for every arm $k\in\KK$ the rewards are distributed according to the C.D.F. $F_{k}(\mu)$. Let assume w.l.o.g. that $\mu^{*}_{1}=\maxmu$. Then, for $N>0$ and by the fact that $(1-\epsilon)^\frac{1}{\epsilon}\leq e^{-1}$ for every $\epsilon\in(0,1]$, for $\UB(N)=\left(\frac{\G-\ln(\delta)}{\CA N}\right)^{1/\beta}$ it follows that
\begin{equation}\label{eq:delta:bound:pre}
P\left(V^{1}_{N}\leq\maxmu-\UB(N)\right)=\left(F_{1}\left(\maxmu-\UB(N)\right)\right)^{N}\leq \left(1-\CA\left(\UB(N)\right)^{\beta}\right)^{N}\leq \delta e^{-L},
\end{equation}
where $V^{k}_{N}$ is the largest reward observed from arm $k\in\KK$ after this arm has been sampled for $N$ times. Hence, at every time step $t$, by the definition of $Y^{1}_{C^{t}(1)}$ and Equations \eqref{time:bound} and \eqref{eq:delta:bound:pre}, by applying the union bound, it follows that
\begin{equation}\label{eq:delta:bound}
P\left(Y^{1}_{C^{t}(1)}\leq\maxmu\right)\leq P\left(V^{1}_{C^{t}(1)}\leq\maxmu-\UB(C^{t}(1))\right)\leq \sum_{t=1}^{\exp\left(\frac{\G}{3}\right)}P\left(V^{1}_{N}\leq\maxmu-\UB(N)\right)\leq \delta e^{-\frac{2L}{3}}.
\end{equation}

Since by the condition in step \ref{cond2}, it is obtained that when the algorithm stops
\begin{equation*}
V^{k^{*}}_{C^{t}(k^{*})}>Y^{k^{*}}_{C^{t}(k^{*})}-\epsilon,
\end{equation*}
and by the fact that for every time step
\begin{equation*}
Y^{k^{*}}_{C^{t}(k^{*})}\geq Y^{1}_{C^{t}(1)},
\end{equation*}
it follows by Equation \eqref{eq:delta:bound} that
\begin{equation*}
P\left(V^{k^{*}}_{C^{t}(k^{*})}\leq\maxmu-\epsilon\right)\leq P\left(Y^{1}_{C^{t}(1)}\leq\maxmu\right)\leq\delta e^{-\frac{2L}{3}}.
\end{equation*}
Therefore, it follows that the algorithm returns a reward greater than $\maxmu-\epsilon$ with a probability larger than $1-\delta$. So, it is $(\epsilon,\delta)$-correct.\\

For proving the bound on the expected sample complexity of the algorithm we define the following sets:
$$M(\epsilon)=\{l\in\KK|\maxmu-\mu^{*}_{l}<\epsilon\},\quad N(\epsilon)=\{l\in\KK|\maxmu-\mu^{*}_{l}\geq\epsilon\}.$$
As before, we assume w.l.o.g. that $\mu^{*}_{1}=\maxmu$.
For the case in which
\begin{equation*}
E_{1}\triangleq\bigcap_{1\leq t<T}\left\{Y^{1}_{C^{t}(1)}\geq\maxmu\right\},
\end{equation*}
occurs, since $V^{k}_{C^{t}(k)}\leq\mu^{*}_{k}$ for every $k\in\KK$, and every time step, it follows that the necessary condition for sampling from arm $k$,
\begin{equation*}
Y^{k}_{C^{k}(1)}\geq Y^{1}_{C^{t}(1)},
\end{equation*}
occurs only when the event
\begin{equation*}
E_{2}(t)\triangleq\left\{\mu^{*}_{k}+\UB\left(C^{t}(k)\right)\geq \maxmu\right\},
\end{equation*}
occurs. But
\begin{equation*}
E_{2}(t)\subseteq \left\{C^{t}(k)\leq \frac{\G-\ln(\delta)}{\CA \left(\maxmu-\mu^{*}_{k}\right)^{\beta}}\right\}.
\end{equation*}
Therefore, it is obtained that
\begin{equation}\label{bound_set_N}
C^{T}(k)\leq \lfloor\frac{\G-\ln(\delta)}{\CA \left(\maxmu-\mu^{*}_{k}\right)^{\beta}}\rfloor+1.
\end{equation}
By using the bound in Equation \eqref{C_bound} for the arms in the set $M(\epsilon)$, the bound in Equation \eqref{bound_set_N} for the arms in the set $N(\epsilon)$ and the bound in Equation \eqref{time:bound}, it is obtained that
\begin{equation}\label{eq:final:1}
E[T]\leq \left(1-P\left(E_{1}\right)\right)e^{\frac{\G}{3}}+P\left(E_{1}\right)\Phi\left(\epsilon\right),
\end{equation}
where
\begin{equation*}
\Phi\left(\epsilon\right)\triangleq\left(\sum_{k\in N(\epsilon)}\left(\lfloor\frac{\G-\ln(\delta)}{\CA \left(\maxmu-\mu^{*}_{k}\right)^{\beta}}\rfloor+1\right)+\sum_{k\in M(\epsilon)}\left(\lfloor\frac{\G-\ln(\delta)}{\CA \epsilon^{\beta}}\rfloor+1\right)\right).
\end{equation*}
In addition, by Equation \eqref{eq:delta:bound}, the bound in Equation \eqref{time:bound} and by applying the union bound, it follows that
\begin{equation*}
P\left(E_{1}\right)\geq1-\sum_{t=1}^{T}P\left(Y^{1}_{C^{t}(1)}<\maxmu\right)\geq 1-\delta e^{-\frac{2L}{3}}e^{\frac{L}{3}}=1-\delta e^{-\frac{L}{3}}.
\end{equation*}
So,
\begin{equation}\label{eq:final:2}
1-P\left(E_{1}\right)\leq\delta e^{-\frac{L}{3}}.
\end{equation}
Furthermore, by the definitions of the sets $N(\epsilon)$ and $M(\epsilon)$, it can be obtained that
\begin{equation}\label{eq:final:3}
\Phi\left(\epsilon\right)\leq\sum_{k\in \KK}\lfloor\frac{\G-\ln(\delta)}{\CA \left(\max\left(\epsilon,\maxmu-\mu^{*}_{k}\right)\right)^{\beta}}\rfloor+1.
\end{equation}

Therefore, by Equation \eqref{eq:final:1}, \eqref{eq:final:2} and \eqref{eq:final:3} the bound on the sample complexity is obtained.

\qed
\end{proof*}

\subsection{Maximal Eliminator}
The algorithm starts by sampling a certain number of times from each arm. Then, it repeatedly calculates an index for each arm which can be interpreted as a certain upper bound on the maximal reward of this arm, and eliminates arms for which that index is below the maximal sampled reward so far. Then it sample from only the retained arms (those arms which have not been eliminated) a number of times that is doubled at each sampling phase. This idea is similar to that in the Median Elimination Algorithm provided in \cite{Dar}.
\begin{algorithm}[tb]
	\caption{Maximal Eliminator (ME) Algorithm}
	\label{alg:eliminator}
	\begin{algorithmic}[1]
	\STATE {\bfseries Input:} Model parameters $\epsilon_{0}>0$, $\CA>0$ and $\beta\geq0$, constants $\delta>0$ and $\epsilon>0$.\\ Define $\GME=\ln\left(12\ln\left(|\KK|\left(1+\frac{-\ln(\delta)}{\CA\epsilon^{\beta}}\right)\right)\right)$.
		\STATE {\bfseries Initialization:} A set of arms $\KK_{t=1}=\KK$ and counter $t=1$. 
		\STATE Sample $N_{t}$ times from each arm in the set $\KK_{t}$, where $N_{t}=2^{t-1}\left(\lfloor\frac{\GME-\ln(\delta)}{\CA\epsilon^{\beta}_{0}}\rfloor+1\right)$.\label{begin:mu:eliminator}
		\STATE Compute $Y^{k}=V^{k}+\UB(N_{t+1}-N_{0})$,
		\\where $V^{k}$ is the largest reward observed from arm $k$ and $\UB(N)=\left(\frac{\GME-\ln(\delta)}{\CA N}\right)^{1/\beta}$.
            \label{cond1:eliminator}
		\STATE If $\UB(N_{t+1}-N_{0})<\epsilon$, stop and return the largest sampled reward.\\
		Else, set $t=t+1$, $\KK_{t}=\{k\in\KK_{t-1}|Y^{k}\geq\max_{j\in\KK_{t-1}}V^{j}\}$ and return to step \ref{begin:mu:eliminator}.\label{cond2:eliminator}
	\end{algorithmic}
\end{algorithm}

We do not provide performance analysis for Algorithm \ref{alg:eliminator}. However, since the number of times at which the confidence bounds should be correct (times at which the algorithm eliminates arms) is only logarithmic in the number of total samples, we have $\GME=\ln(2\G)$ (where $\G$ is defined in Algorithm \ref{alg:example} and the factor $2$ arises because of the doubling). Therefore, we believe that the upper bound on the sample complexity of Algorithm \ref{alg:eliminator} would be that of Algorithm \ref{alg:example} multiplied by $\frac{2\GME}{\G}$. So, the upper bound would be of the same order of the lower bound in Theorem \ref{thm:lower:bound} up to double logarithmic terms.

\section{Comparison with The Unified-Arm Model}\label{sec:comp}
In this section, we analyze the improvement in the sample complexity obtained by utilizing the multi arm property (the ability to choose from which arm to sample at each time step) compared to a model in which all the arms are unified into a unified arm, so that the sample is effectively obtained from a random arm. 
In the unified-arm model, when the agent samples from this unified arm, a certain arm (among the multi arm) is chosen uniformly and a reward is sampled from this arm. We denote the CDF of the unified arm as $F(\mu)$, with $ F=\frac{1}{|K|} \sum_{k\in \KK} F_k $.
By Assumption \ref{assumption:1}, $1-F(\mu)\geq\frac{\CA\left(\maxmu-\mu\right)^{\beta}}{|\KK|}$, and the corresponding maximal reward is $\maxmu$.


In the remainder of this section, we provide a lower bound on the sample complexity and an $(\epsilon,\delta)$-correct algorithm that attains the same order of this bound for the unified-arm model. (Note that the 
lower bound in Theorem \ref{thm:lower:bound} is meaningless for $|K|=1$.)
Then, we discuss which approach (multi-arm or unified-arm) is better for different model parameters, 
and provide examples that illustrate these cases.

\subsection{Lower Bound}
The following Theorem provides a lower bound on the sample complexity for the unified-arm model.
\begin{theorem}\label{thm:lower:single}
Suppose $\epsilon_{0}\leq\left(\frac{|\KK|}{2\CA}\right)^{\frac{1}{\beta}}$, $\beta\leq1$ and let $\epsilon\in(0,\epsilon_{0})$, $\delta\in(0,1)$. Then, under Assumption \ref{assumption:1}, for every $(\epsilon,\delta)$-correct algorithm, it holds that
\begin{equation}
E[T]\geq \frac{|\KK|}{4\CA\epsilon^{\beta}}\ln\left(\frac{3}{5\delta}\right).
\end{equation}
\end{theorem}
The proof is provided in Appendix B and is based on the a similar idea to that of Theorem \ref{thm:lower:bound}.

\subsection{Algorithm}
In Algorithm \ref{alg:example_S} a certain number of rewards is sampled, and the algorithm chooses the best one among them. In the following Theorem we provide a bound on the sample complexity achieved by Algorithm \ref{alg:example_S}.
\begin{algorithm}[tb]
	\caption{Unified-Arm Algorithm}
	\label{alg:example_S}
	\begin{algorithmic}[1]
		\STATE {\bfseries Input:} Constants $\delta>0$, $\epsilon>0$.
		\STATE Sample $\lceil\frac{-\ln(\delta)|\KK|}{\CA\epsilon^{\beta}}\rceil+1$ arms from the arm.\label{begin:mu_S}
		\STATE Return the best sampled arm.
	\end{algorithmic}
\end{algorithm}

\begin{theorem}\label{thm:alg:Single}
Under Assumption \ref{assumption:1}, Algorithm \ref{alg:example_S} is $(\epsilon,\delta)$-correct, with a sample complexity bound of
\begin{equation*}
E[T]\leq
\frac{|\KK|\ln(\delta^{-1})}{\CA\epsilon^{\beta}}+2.
\end{equation*}
\end{theorem}
The proof is provided in Appendix C. Note that the upper bound on the sample complexity is of the same order as the lower bound in Theorem \ref{thm:lower:single}.

\subsection{Comparison and Examples}
 To find when the multi-arm algorithm is helpful, we can compare the upper bound on the sample complexity provided in Theorem \ref{thm:alg1} for Algorithm \ref{alg:example} (multi-arm case) with the lower bound for the unified-arm model in Theorem \ref{thm:lower:single}.
 
{\em Case 1:} Suppose first that arm 1 is best: $\mu_1^*=\maxmu$, while all the other arms fall short significantly 
compared to the required accuracy $\epsilon$: $\mu_k^*  \ll \maxmu - \epsilon$, for $k\neq 1$.
\\In this case $\frac{1}{\epsilon^{\beta}}\gg\frac{1}{\left(\max\left(\epsilon,\maxmu-\mu^{*}_{k}\right)\right)^{\beta}}$, for $k\neq 1$. Hence the upper bound on sample complexity of Algorithm \ref{alg:example} (multi-arm case) will be smaller than the lower bound for the unified-arm model in Theorem \ref{thm:lower:single}. We now provide an example which illustrate case 1 numerically.
\begin{example}[Case 1]\label{ex:1}
	Let $|\KK|=10^4$, $\mu^{*}_{1}=0.9$, $\mu^{*}_{k}=0.1\ \forall k\in\KK\setminus\{1\}$, $\beta=1$ and $\CA=0.01$. 
 	For $\epsilon=10^{-4}$ and $\delta=10^{-3}$ the sample complexity attained by Algorithm \ref{alg:example} is $3.52\times10^8$.  The lower bound for the unified-arm model is $1.59\times10^{10}$. The sample complexity attained by Algorithm \ref{alg:example_S} (for the unified-arm model model) is $6.9\times10^{10}$.
\end{example}
{\em Case 2:} Consider next the opposite case, where there are many optimal arms and few that are worse: 
say $\mu_1^* \ll \maxmu -\epsilon$, while $\mu_k^*=\maxmu$ for all $k\neq 1$.
\\In this case $\frac{1}{\epsilon^{\beta}}=\frac{1}{\left(\max\left(\epsilon,\maxmu-\mu^{*}_{k}\right)\right)^{\beta}}$, for $k\neq 1$. Hence, since there is a logarithmic-in-$\frac{|\KK|}{\epsilon}$ multiplicative factor in the upper bound on the sample complexity of Algorithm \ref{alg:example} (multi-arm case), this bound will be larger than the lower bound for the unified-arm model in Theorem \ref{thm:lower:single}. The following example illustrate case 2 numerically.
\begin{example}[Case 2]\label{example:2}
 	Let $|\KK|$, $\CA$, $\beta$, $\delta$ and $\epsilon$ remain the same as in Example \ref{ex:1}, and let $\mu^{*}_{1}=0.1$ and $\mu^{*}_{k}=0.9\ \forall k\in\KK\setminus\{1\}$. The sample complexity of Algorithm \ref{alg:example} is $1.56\times10^{12}$, which is larger than the sample complexity of Algorithm \ref{alg:example_S} which is $6.9\times10^{10}$.
\end{example}

As shown in Example \ref{example:2}, in some cases the bound on the sample complexity of Algorithm \ref{alg:example} (multi-arm) is larger than that of Algorithm \ref{alg:example_S} (unified-arm). By comparing the upper bounds of these algorithms, we believe that the logarithmic in $\frac{\KK}{\epsilon}$ factor in the bound of Algorithm \ref{alg:example} may not be required.

As observed by comparing the lower and upper bounds for the multi-arm and the unified-arm model, the unified-arm algorithm provides a tighter upper bound (compared to the matching lower bound). Therefore, when the benefit obtained by the multi-arm model is small (i.e., when there are a lot of good arms) the profit obtained by applying the multi-arm Algorithm turns out to be loss.


\section{Conclusion}\label{conc}

In this paper we have developed corresponding lower and upper bounds on the sample complexity, 
which are essentially the same order up to a logarithmic term in $\frac{|\KK|}{\epsilon}$ for the Max $K$-Armed Bandit problem.

These results were compared to the unified-arm model, where the learning algorithm effectively unifies the different arms into one. While the multi-arm algorithm usually performs better, in some cases, in particular when most arms
are optimal, the unified arm algorithm may provide better performance. It still remains to be shown whether an 
algorithm that provides the performance benefits of both approaches may be devised. 
 
Another direction for future work concerns the relaxation or generalization of our Assumption 1, which requires a known
lower bound on the tail distribution of the rewards. 

\vskip 0.2in
\bibliographystyle{ieeetr} 
\bibliography{sample_new}
\newpage
\section{Appendix A}\label{sec:proof:lowerbound}
\begin{proof*}[Theorem \ref{thm:lower:bound}]
Let $\overline{\mu}_{k}=\sup_{\mu\in\mathbb{R}}\{\mu|F_{k}(\mu)\leq1-A\epsilon_{0}^{\beta}\}$ for every $k\in\KK$. Then, we define the following set of hypotheses $\{H_0,H_1,\dots,H_{|K|}\}$:
	\begin{equation*}
	H_{0}:\quad f^{H_{0}}_{k}(\mu)=f_{k}(\mu)\quad \forall k\in\KK,
	\end{equation*}
	and, for every $k=1,\dots,|K|$,  
	\begin{align*}
	H_{k}:
	\quad & f^{H_{k}}_{l}(\mu)=f_{l}(\mu),\quad l\neq k,\\
	& \text{if $\epsilon_{0}\leq \left(\maxmu-\mu^{*}_{k}+\epsilon\right)$:}\quad  f^{H_{k}}_{k}(\mu)=\gamma^{1}_{k}f_{k}(\mu)\boldsymbol{1}_{(-\infty,\mu^{*}_{k})}(\mu)+A\beta\left(\maxmu+\epsilon-\mu\right)^{\beta-1}\boldsymbol{1}_{\left(\maxmu+\epsilon-\epsilon_{0},\maxmu+\epsilon\right]}(\mu),\\
	&\text{if $\epsilon_{0}>\left( \maxmu-\mu^{*}_{k}+\epsilon\right)$ :}\quad
		\begin{aligned} f^{H_{k}}_{k}(\mu)=&\gamma^{2}_{k}f_{k}(\mu)\boldsymbol{1}_{(-\infty,\overline{\mu}_{k})}(\mu)+\gamma^{3}_{k}f_{k}(\mu)\boldsymbol{1}(\mu=\overline{\mu}_{k})+f_{k}(\mu)\boldsymbol{1}_{(\overline{\mu}_{k},\mu^{*}_{k}]}(\mu)
		\\&+A\beta\left(\maxmu+\epsilon-\mu\right)^{\beta-1}\boldsymbol{1}_{\left(\mu^{*}_{k},\maxmu+\epsilon\right]}(\mu)
		\end{aligned},\\
		\end{align*}
	where $f_{k}(\mu)$ is the probability density function of arm $k\in\KK$, $\boldsymbol{1}_\Theta$ stand for the indicator function of the set $\Theta$, $\gamma^{1}_{k}=1-A\epsilon_{0}^{\beta}$, $\gamma^{2}_{k}=1-\frac{A\left(\maxmu-\mu^{*}_{k}+\epsilon\right)^{\beta}}{1-A\epsilon_{0}^{\beta}}$ and $\gamma^{3}_{k}$ is chosen such that $\int_{0}^{1}f^{H_{k}}_{k}(\mu)d\mu=1$.
	
	Note that since for every $x_{1},x_{2}\geq0$ it follows that $x_{1}^{\beta}+x_{2}^{\beta}\geq \left(x_{1}+x_{2}\right)^{\beta}$ for $\beta\leq1$, Assumption \ref{assumption:1} holds for hypotheses $\{H_1,\dots,H_{|K|}\}$.
	
	To further bound $\gamma^{2}_k$ and $\gamma^{3}_k$, note that since $\epsilon_{0}\leq\left(4\CA\right)^{-1/\beta}$,
	$$1-2\CA\left(\maxmu-\mu^{*}_{k}+\epsilon\right)^{\beta}\leq\gamma^{2}_k\leq1\,.
	$$
	Let $P_{k}$ stands for the mass of an atom in the probability function of arm $k\in\KK$ at the point $\overline{\mu}_{k}$ (if there is one), then we note that
	$$
	1=\int_{-\infty}^{\infty}f^{H_{k}}_{k}(\mu) d\mu= \gamma^{2}_{k}\left(F_{k}(\overline{\mu}_{k})-P_{k}\right)+\gamma^{3}_{k}P_{k}+1-F_{k}(\overline{\mu}_{k})+A\left(\maxmu-\mu^{*}_{k}+\epsilon\right)^{\beta}\triangleq\Phi(\gamma^{2}_{k},\gamma^{3}_{k})\,,
	$$
	but, since $F_{k}(\overline{\mu}_{k})\geq1-A\epsilon_{0}^{\beta}$, for $\gamma^{2}_{k}=\gamma^{3}_{k}$ it follows that $\Phi(\gamma^{2}_{k},\gamma^{3}_{k})\leq1$.
	So, since $\Phi(\gamma^{2}_{k},\gamma^{3}_{k})$ increases in $\gamma^{3}_{k}$ it is obtained that $\gamma^{3}_{k}\geq\gamma^{2}_{k}$. Finally, it follows that in the case of $\epsilon_{0}\leq \left(\maxmu-\mu^{*}_{k}+\epsilon\right)$,
	$$\gamma_{k}\leq\gamma^{1}_{k}\,,$$
	and in the case of $\epsilon_{0}> \left(\maxmu-\mu^{*}_{k}+\epsilon\right)$,
	$$\gamma_{k}\leq\min\left(\gamma^{2}_{k},\gamma^{3}_{k}\right)\,,$$
	where
	$$\gamma_{k}\triangleq1-2A\left(\min\left(\epsilon_{0},\maxmu-\mu^{*}_{k}+\epsilon\right)\right)^{\beta}$$
	
	If hypothesis $H_{k}$ ($k\neq 0$) is true, then $\mu_k^* \geq \mu_l^* +\epsilon$  for all $l\neq k$, hence the algorithm should provide a reward from arm $k$ with probability larger than $1-\delta$. We use $E^{H}_{k}$ and $P^{H}_{k}$ to denote the expectation and probability, respectively, under the algorithm being considered and hypothesis $H_{k}$. Further, for every $k\in\KK$ let
	\begin{equation*}
	t_{k}=\frac{1}{4\left(1-\gamma_{k}\right)}\ln\left(\frac{3}{16\delta}\right),
	\end{equation*}
	and let $T_{k}$ stands for the number of samples from arm $k$.
	
	Suppose now that our algorithm is $(\epsilon,\delta)$-correct under $H_{0}$, and that $E^{H}_0[T_{k}]\leq t_{k}$ for
	some $k\in\KK$. 
	We will show that this algorithm cannot be $(\epsilon,\delta)$-correct under hypothesis $H_{k}$. Therefore, an $(\epsilon,\delta)$-correct algorithm must have $E^{H}_0[T_{k}]>t_{k}$ for all $k\in\KK$.
	
	Define the following events:
	\begin{itemize}
		\item $A_{k}=\{T_{k}\leq 4t_{k}\}$. It easily follows from
		$
		4t_{k}\left(1-P^{H}_{0}(A_{k})\right)\leq E^{H}_{0}[T_{k}]
		$
		that if $E^{H}_{0}[T_{k}]\leq t_{k}$, then $P^{H}_{0}(A_{k})\geq\frac{3}{4}$.
		\item
		Let $B_{k}$ stand for the event under which the chosen arm at termination is $k$, and $B^{C}_{k}$ for its complement. Since $P^{H}_{0}\left(B_{k'}\right)>\frac{1}{2}$ can hold for one arm at most, it follows that $P^{H}_{0}\left(B^{C}_{k}\right)>\frac{1}{2}$ for every $k\in \KK\setminus\{k'\}$ for some $k'$.
		\item
		Let $C_{k}$ to be the event under which all the samples obtained from arm $k$ are on the interval $(-\infty,\mu^{*}_{k}]$. Clearly, $P^{H}_{0}(C_{k})=1$.
	\end{itemize}
	Define now the intersection event $S_{k}=A_{k}\cap B_{k}^{C}\cap C_{k}$. We have just shown that for every $k\in\KK\setminus\{k'\}$ it holds that $P^{H}_{0}(A_{k})\geq\frac{3}{4}$, $P^{H}_{0}(B_{k}^{C})>\frac{1}{2}$ and $P^{H}_{0}(C_{k})=1$, from which it follows that $P^{H}_{0}\left(S_{k}\right)>\frac{1}{4}$ for $k\neq k'$.
	Further, observe that for every history $h_N$ of $N$ samples for which the event $C_{k}$ holds, it holds that $\frac{dP^{H}_{k}}{dP^{H}_{0}}(h_N) \geq(\gamma_{k})^{N}$. We therefore obtain the following inequalities,
	\begin{align*}
	P^{H}_{k}\left(B^{C}_{k}\right) &\geq P^{H}_{k}\left(S_{k}\right)=E^{H}_{0}\left[\frac{dP^{H}_{k}}{dP^{H}_{0}}\Indi{S_{k}}\right]
	\geq \gamma_{k}^{-4t_{k}}P^{H}_{0}\left(\Indi{S_{k}}\right) \\
	& >\frac{1}{4}\gamma_{k}^{-4t_{k}}\geq \frac{1}{4}e^{-\ln\frac{3}{16\delta}}\geq\delta,
	\end{align*}
	where in the last inequality we used the facts that $\left(1-\epsilon\right)^{\frac{1}{\epsilon}}\geq e^{-1}$.
	
	We found that if an algorithm is $(\epsilon,\delta)$-correct under hypothesis $H_{0}$ and $E_{0}[T_{k}]\leq t_{k}$ for some $k\neq k'$, then, under hypothesis $H_{k}$ this algorithm returns a sample that is smaller by at least $\epsilon$ than the maximal possible reward with probability of $\delta$ or more, hence the algorithm is not $(\epsilon,\delta)$-correct.
	Therefore, any $(\epsilon,\delta)$-correct algorithm must satisfy $E_{0}[T_{k}]> t_{k}$ for all of arms except possibly for one (namely, for the one $k'$ for which $P_{0}\left(B^{C}_{k'}\right)\leq\frac{1}{2}$). In addition $t_{k^{*}}\geq t_{k'}$, where $k^{*}$ is the optimal arm (namely, $\mu^{*}_{k^{*}}=\maxmu$). Hence the lower bound is obtained.
	
	\qed
\end{proof*}

\section{Appendix B}
\begin{proof*}[Theorem \ref{thm:lower:single}]
First , we define the following hypotheses:\\
\begin{equation*}
H_{0}:\quad f^{H_{0}}(\mu)=f(\mu),
\end{equation*}
and
\begin{equation*}
H_{1}:
\quad f^{H_{1}}(\mu)=\gamma f(\mu)+\frac{\CA}{|K|}\beta\left(\maxmu+\epsilon-\mu\right)^{\beta-1}\boldsymbol{1}_{\left(\maxmu,\maxmu+\epsilon\right]}(\mu),
\end{equation*}
where, as in the proof of Theorem \ref{thm:lower:bound}, $f(\mu)$ is the probability density function of the unified arm, $\boldsymbol{1}_A$ stand for the indicator function of the set $A$,  and $\gamma$ is chosen such that $\int_{0}^{1}f^{H_{1}}(\mu)d\mu=1$.

Note that since for every $x_{1},x_{2}\geq0$ it follows that $x_{1}^{\beta}+x_{2}^{\beta}\geq \left(x_{1}+x_{2}\right)^{\beta}$ for $\beta\leq1$, Assumption \ref{assumption:1} holds for hypothesis $H_{1}$.

To further bound $\gamma$, note that 
\begin{equation*}
1=\int_{-\infty}^{\infty}f^{H_{1}}(\mu) d\mu  = \gamma  +\frac{\CA\epsilon^{\beta}}{|\KK|},
\end{equation*}
Therefore,
\begin{equation*}
\gamma=1-\frac{\CA\epsilon^{\beta}}{|\KK|}.
\end{equation*}
If hypothesis $H_{1}$ is true, the algorithm should provide a reward greater than $\maxmu$. We use $E_{l}$ and $P_{l}$ (where $l\in\{0,1\}$) to denote the expectation and probability respectively, under the
algorithm being considered and under hypothesis $H_{l}$.
Now, let
\begin{equation*}
t=\frac{1}{4\left(1-\gamma\right)}\ln\left(\frac{3}{5\delta}\right),
\end{equation*}
and recall that $T$ stands for the total number of samples from the arm.

Now, we assume we run an algorithm which is $(\epsilon,\delta)$-correct under $H_{0}$ and that $E_{0}[T]\leq t$ for this algorithm. We will show that this algorithm cannot be $(\epsilon,\delta)$-correct under hypothesis $H_{1}$. Therefore, an $(\epsilon,\delta)$-correct algorithm must have $E_{0}[T]>t$.

Define the following events:
\begin{itemize}
	\item
$A=\{T\leq 4t\}$. By the same consideration as in the proof of Theorem \ref{thm:lower:bound} (for the events $\{A_{k}\}_{k\in\KK}$), it follows that if $E_{0}[T]\leq t$, then $P_{0}(A)\geq\frac{3}{4}$.
\item
Let $B$ stand for the event under which the chosen sample is smaller or equal to $\maxmu$, and $B^{C}$ for its complementary. Clearly, $P_{0}\left(B\right)=1$.
\item
We define the event $C$ to be the event under which all the samples obtained from the unified arm are on the interval $[-\infty,\maxmu]$. Clearly, $P_{0}(C)=1$.
\end{itemize}

Define now the intersection event $S=A\cap B^{C}\cap C$. We have shown that $P_{0}(A)\geq\frac{3}{4}$, $P_{0}(B)=1$ and $P_{0}(C)=1$, from which it is obtained that $P_{0}\left(S\right)\geq\frac{3}{4}$.
In addition, since for every history $h_{N}$ of $N$ samples, for which the event $C$ holds, it is obtained that $\frac{dP_{1}}{dP_{0}}\left(h_{N}\right)\geq\gamma^{N}$, we have the following,
\begin{equation*}
\begin{aligned}
P_{1}\left(B\right)&\geq P_{1}\left(S\right)=E_{0}\left[\frac{dP_{1}}{dP_{0}}\Indi{S}\right]
\geq \gamma^{-4t}P_{0}\left(\Indi{S}\right)
\\&\geq\frac{3}{4}\gamma^{-4t}\geq \frac{3}{4}e^{-\ln\frac{3}{5\delta}}\geq\delta,
\end{aligned}
\end{equation*}
where in the last inequality we used the facts that $\left(1-\epsilon\right)^{\frac{1}{\epsilon}}\geq e^{-1}$.

We found that if an algorithm is $(\epsilon,\delta)$-correct under hypothesis $H_{0}$ and $E_{0}[T]\leq t$, then, under hypothesis $H_{1}$ this algorithm returns a sample that is smaller by at least $\epsilon$ than the maximal possible reward with a probability of $\delta$ or more, hence the algorithm is not $(\epsilon,\delta)$-correct.
Therefore, any $(\epsilon,\delta)$-correct algorithm, must satisfy $E_{0}[T]> t$. Hence the lower bound is obtained.

\qed
\end{proof*}

\section{Appendix C}
\begin{proof*}[Theorem \ref{thm:alg:Single}]
Since sampling from the unified arm consists of choosing one arm out of the $|\KK|$ arms (with equal probability), and then, sampling from this arm, it follows that, $F\left(\maxmu-\epsilon\right)\leq\left(1-\frac{\CA\epsilon^{\beta}}{|\KK|}\right)$. Also, we note that $(1-\epsilon)^\frac{1}{\epsilon}\leq e^{-1}$ for every $\epsilon\in(0,1]$. Therefore, for $N=\lceil\frac{-\ln(\delta)|\KK|}{\CA\epsilon^{\beta}}\rceil+1$,
\begin{equation}
P\left(V^{1}_{N}<\maxmu-\epsilon\right)=\left(F\left(\maxmu-\epsilon\right)\right)^{N}\leq \left(1-\frac{\CA\epsilon^{\beta}}{|\KK|}\right)^{N}< \delta,
\end{equation}
where $V^{1}_{N}$ is the largest reward observed among the first $N$ samples. Hence, the algorithm is $(\epsilon,\delta)$-correct. The bound on the sample complexity is immediate from the definition of the algorithm.

\qed
\end{proof*}
\end{document}